\documentclass[a4paper,conference]{IEEEtran}

\ifCLASSINFOpdf
\else
\fi

\usepackage{hyperref}       
\usepackage{url}            
\usepackage{booktabs}       
\usepackage{nicefrac}       
\usepackage{microtype}      

\usepackage{epsfig}
\usepackage{graphicx}
\usepackage{amsmath}
\usepackage{amssymb}
\usepackage{amsfonts,amsthm,bm}
\usepackage{graphics}
\usepackage{float}
\usepackage{array}
\usepackage{booktabs}
\usepackage{multirow}
\usepackage[table,xcdraw]{xcolor}
\usepackage{lipsum}
\usepackage{hhline}
\usepackage{makecell}
\usepackage[numbers,sort,compress]{natbib}

\usepackage{caption, subcaption}

\newcommand{\muv}{{\bm{\mu}}}

\newcommand{\zv}{\mathbf{z}}
\newcommand{\xv}{\mathbf{x}}

\newcommand{\nv}{\mathbf{n}}

\newcommand{\Sigmam}{\bm{\Sigma}}

\newcommand{\Xm}{\mathbf{X}}
\newcommand{\Zm}{\mathbf{Z}}

\newcommand{\PreserveBackslash}[1]{\let\temp=\\#1\let\\=\temp}
\newcolumntype{C}[1]{>{\PreserveBackslash\centering}p{#1}}
\newcolumntype{R}[1]{>{\PreserveBackslash\raggedleft}p{#1}}
\newcolumntype{L}[1]{>{\PreserveBackslash\raggedright}p{#1}}

\begin{document}
\title{Beyond cross-entropy: learning highly separable feature distributions for robust and accurate classification}

\author{
\IEEEauthorblockN{Arslan Ali, Andrea Migliorati, Tiziano Bianchi, Enrico Magli}
\IEEEauthorblockA{Department of Electronics and Telecommunications\\
Politecnico di Torino (Turin, Italy)\\
\texttt{\textit{name.surname}@polito.it}}
}

\maketitle

\begin{abstract}
Deep learning has shown outstanding performance in several applications including image classification. However, deep classifiers are known to be highly vulnerable to adversarial attacks, in that a minor perturbation of the input can easily lead to an error. Providing robustness to adversarial attacks is a very challenging task especially in problems involving a large number of classes, as it typically comes at the expense of an accuracy decrease. In this work, we propose the Gaussian class-conditional simplex (GCCS) loss: a novel approach for training deep robust multiclass classifiers that provides adversarial robustness while at the same time achieving or even surpassing the classification accuracy of state-of-the-art methods. Differently from other frameworks, the proposed method learns a mapping of the input classes onto target distributions in a latent space such that the classes are linearly separable. Instead of maximizing the likelihood of target labels for individual samples, our objective function pushes the network to produce feature distributions yielding high inter-class separation. The mean values of the distributions are centered on the vertices of a simplex such that each class is at the same distance from every other class. We show that the regularization of the latent space based on our approach yields excellent classification accuracy and inherently provides robustness to multiple adversarial attacks, both targeted and untargeted, outperforming state-of-the-art approaches over challenging datasets. 
\end{abstract}

\section{Introduction}
\label{sec:intro}

\begin{figure*}[t!]
\centering

  \centering
  \includegraphics[width=0.75\linewidth]{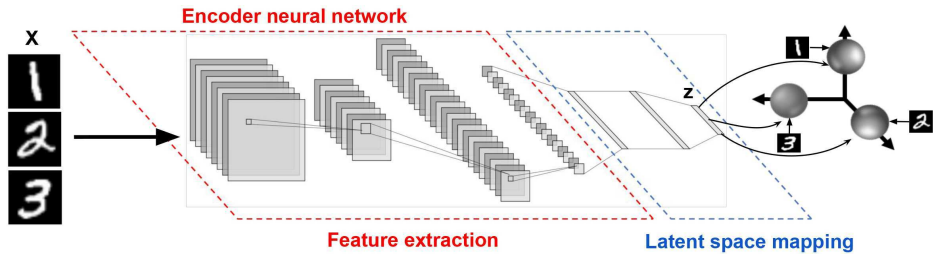}
\caption[]{The GCCS architecture takes input data and learns discriminative features that are mapped onto Gaussian target distributions in the latent space.}
\label{fig:RobustNet-arch}
\vspace*{-0.5cm}
\end{figure*}

In recent years, deep neural networks have reached accuracy comparable with or even greater than that of humans in visual tasks such as recognizing traffic signs \cite{cirecsan2012multi}, handwritten digits \cite{wan2013regularization}, and faces \cite{sun2014deep}. Also, they have shown excellent performance at learning complex mappings \cite{almahairi2018augmented} and addressing difficult classification tasks \cite{krizhevsky2012imagenet}. However, as their integration in contemporary society grows, they become ever more subject to the action of malicious \textit{adversaries}.

In fact, despite the success of deep neural networks, many obstacles still hinder their use in fields where security is essential, such as systems for autonomous driving and medical diagnostics \cite{eykholt2018robust,finlayson2019adversarial}. A major threat is represented by adversarial perturbations, a set of techniques used to tamper with the inputs of a neural network. The modifications are often invisible to the human eye but may still be able to disrupt the algorithm operation and cause unexpected, undesired outputs. Malicious attackers could exploit such fallacies to cause malfunctions in systems, and the attack would be very hard to detect. 

Although many countermeasures have been proposed, an effective defence mechanism against the broad spectrum of adversarial perturbations is not available yet. In particular, a downside of deep learning techniques is that the learned decision boundaries in the feature space are highly complex and non-linear \cite{fawzi2017classification}. Works addressing this specific problem \cite{fawzi2017classification,robustnessmagazine2017fawzi} concluded that most of the mass of the data points gathers close to the decision boundaries and this may strongly affect the robustness of the classifier against perturbations. Recent techniques tackling this problem can be found in \cite{summers2019improved} and \cite{moosavi2019robustness}, where logit regularization and curvature regularization methods are deployed as adversarial defense respectively, and also in \cite{carmon2019unlabeled} and \cite{pinot2019theoretical}, where theoretical insight is given on the effect of the use of unlabeled data and noise injection at inference time, respectively. At the same time, new techniques are also being developed to craft more successful adversarial attacks \cite{zheng2019distributionally}. In the present paper, adversarial training is not considered as it entails the cost of generating and training on a substantial amount of additional input samples; moreover, adversarial training typically provides robustness against a specific type of attack, whereas we are interested in tackling the robustness problem with a more general approach.

In order to improve the robustness of a classifier in the presence of adversarial perturbation of the inputs, we propose a novel classifier design that goes beyond the cross-entropy loss function. The proposed method employs a new objective function enabling learning of features that maximize inter-class separation and decision variables exhibiting simple and well-defined distributions that are linearly separable in the latent space. The proposed objective function provides state-of-the-art classification while at the same time ensures robustness against adversarial attacks as it is. To correctly evaluate the robustness against adversarial examples, we follow the methodological foundations established in \cite{carlini2019evaluating} and \cite{carlini2017adversarial}. 

The resulting classifier employs simple threshold-based decisions in the regularized latent space. This design provides several benefits: on one hand, the accuracy is typically improved with respect to cross-entropy even in the case of no attacks. On the other hand, such classifier exhibits remarkably improved robustness against adversarial attacks; indeed, due to the uniformity of the distributions of the features in the latent space and the lack of a short path towards a neighboring decision region, the attack strength must be much larger in order to generate a misclassification. Finally, the proposed method can be easily applied to an existing pre-trained cross-entropy based classifier, by continuing the training of the features and classification stage using our proposed loss function. 

This paper presents a detailed assessment and analysis of the proposed method in several image classification problems, providing accuracy results on well-known datasets such as MNIST \cite{lecun1998gradient}, FMNIST \cite{xiao2017fashion}, SVHN \cite{netzer2011reading}, as well as more challenging datasets such as CIFAR10 and CIFAR100 \cite{krizhevsky2009learning}. In particular, we show that our loss function is inherently more robust than cross-entropy. We support our claim by following state-of-the-art robustness evaluation frameworks \cite{carlini2019evaluating}. We validate our approach comparing it to state-of-the-art techniques for adversarial robustness and show that GCCS outperforms those methods under both targeted (PGD \cite{madry2017towards}) and untargeted attacks (JSMA \cite{papernot2016limitations}, TGSM \cite{kurakin2016adversarial}). 
\section{Related works}
\label{sec:related}

The concept of adversarial perturbation was first introduced for spam email detection \cite{dalvi2004adversarial,lowd2005adversarial}. In the following years, Szegedy et al. \cite{szegedy2013intriguing} showed how neural networks can easily be tricked into wrong classification if fed with specifically altered inputs produced considering the sign of the loss function gradients with respect to the inputs. In works such as \cite{goodfellow2014explaining,huang2015learning,moosavi2016deepfool} adversarial samples are used in the training phase as a particular form of data augmentation in order to improve robustness. However, such adversarial training does not prevent adversaries to effectively tamper with the final classification stage \cite{carlini2017adversarial}. Rather, it has been proven that universal adversarial perturbations can be crafted so as to induce wrong classification with high probability independently of the used dataset \cite{moosavi2017universal}, and also to generalize well over different network structures \cite{kurakin2016adversarial,liu2016delving,papernot2016distillation}. Recent theoretical works have also demonstrated that the robustness to adversarial attacks for a classification problem is bounded by limits that cannot be escaped by any classifier since they are dependent on the used datasets, the strength of the attack, and the way perturbations are measured \cite{shafahi2018adversarial}. 

The authors in \cite{ross2018improving} investigated how the effectiveness of adversarial attacks can transfer to models other than the targeted one, and they showed that adversarial examples that are generated to fool a specific model are likely to impact all the models that are trained on the same dataset. Also, \cite{kurakin2016adversarial} concluded that adversarial-generated images are misclassified even when printed out on paper and digitally re-acquired, proving that the phenomenon is relevant in both the digital and the physical domains. Further, \cite{sharif2016accessorize} showed that deep learning methods for face recognition may wrongly classify faces when users are wearing ad-hoc designed adversarial glasses. Finally, \cite{brown2017adversarial} described a method for generating image patches to be placed on input target images in order to cause the neural network to output the desired class. This kind of attack is constructed and performed without knowledge of the targeted image, and it potentially allows the adversarial patch to be widely used with malicious intent after it is distributed over the Internet.

Several papers have also investigated defense techniques against attacks. The authors in \cite{ross2018improving} propose the input gradient regularization method, which is employed during the training phase to force the model to have smooth gradients. They hypothesize that a model trained with gradients that exhibits fewer extreme values is more resistant to adversarial perturbations and that its behavior in response to those attacks is also more easily interpretable. Moreover, \cite{hein2017formal} calculates instance-specific lower bounds on the norm of the input perturbation necessary to alter the decision of the classifier, providing a formal characterization of its robustness. The article also introduces the Cross-Lipschitz regularization functional which forces the differences of the classifier functions at the data points to be constant. Jakubovitz et al. \cite{jakubovitz2018improving} instead suggest a low-complexity regularization technique that uses the Frobenius norm of the Jacobian of the network, which is applied to already trained models as post-processing, robustness-improving step. In particular, while not being an active defence method, the proposed GCCS method ensures improved robustness against adversarial perturbations as it is.

If standard approaches focus on learning the classification boundaries, the proposed GCCS approach instead learns a mapping of the input classes onto target distributions in the latent space. Specifically, an encoder maps features of each class onto Gaussian distributions on a simplex for an arbitrary number of classes, maximizing inter-class separability. Other papers also propose to learn a mapping onto a regularized space, such as \cite{makhzani2015adversarial} and \cite{kingma2013auto} that respectively introduce techniques based on adversarial and variational autoencoders. In \cite{stuhlsatz2012feature} Stuhlsatz et al. present an approach to feature extraction that generalizes the classical Linear Discriminant Analysis (LDA) employing neural networks. The authors in \cite{dorfer2015deep} nonlinearly extend LDA by putting it on top of a deep neural network and maximize the eigenvalues of LDA on the last hidden representation. The primary objective of most discriminant analysis methods is dimensionality reduction \cite{ye2010discriminant}. One of the shortcomings of these methods is that they tend to maximize the distance of the classes that are already well separated, at the expense of poorly-separated neighboring classes, leading to a nonhierarchical pattern in terms of inter-class separability. Another relevant work is RegNet \cite{testa2019learning}, a deep learning technique for biometric authentication that deals with the one-vs-all classification problem of separating authorized users from non-authorized ones. This technique regularizes a two-dimensional latent space through a loss function based on a simplified equation for the Kullback–Leibler divergence; however, this approach is not suitable for high-dimensional classification problems as addressed by GCCS. 

\begin{figure}[t]
    \centering
    \includegraphics[width=0.8\columnwidth]{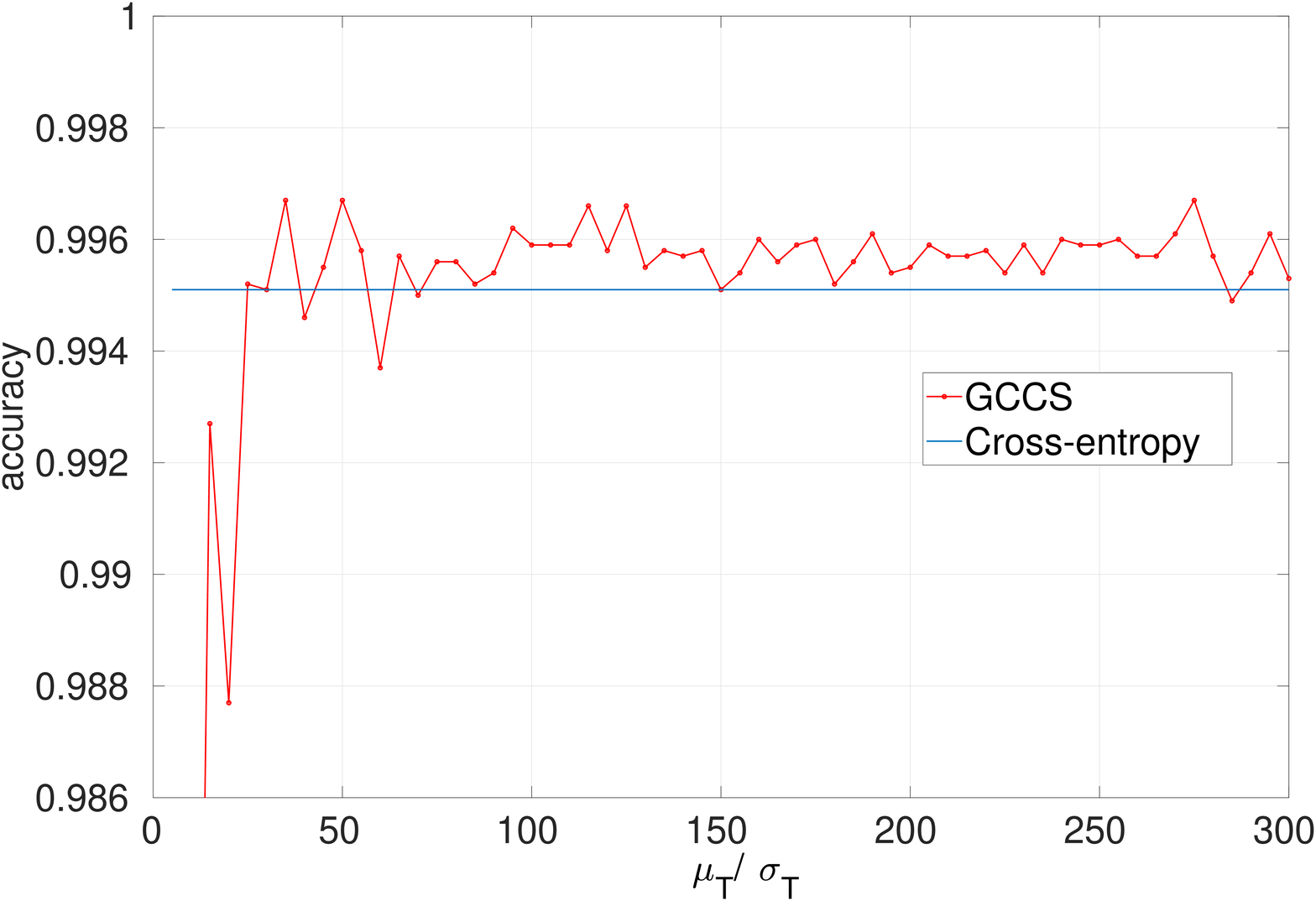}
    \caption[]{Classification accuracy (\%) for GCCS and cross-entropy on MNIST with ResNet-18.}
    \label{fig:parameters_select}
\vspace*{-0.5cm}
\end{figure}

\newcommand{\rulesep}{\unskip\ \vrule\ }
\section{Proposed Method}
\label{sec:proposed}

The proposed method is based on the architecture shown in Fig. \ref{fig:RobustNet-arch}. 
Labeled training data $\Xm$ for a $D$-class classification problem is given as input to a neural network that is composed of a feature extractor and a latent space mapper. The goal of the feature extractor is to learn nonlinear transformations from arbitrary data distributions and extract distinctive and highly separable features. The latent space mapper consists of one or more fully connected layers with the goal of mapping the output $\zv$ onto specific target distributions in a $D$-dimensional latent space (i.e. as many dimensions as the number of classes); no non-linear activation function is employed in the last layer of the mapper. 
It is worth noting that the proposed method does not depend on a specific feature extraction architecture, so existing state-of-the-art architectures can be used for this task.

In order to achieve the desired target, the proposed method needs to define three main components: a target model for the distribution of features in the latent space; a loss function to achieve that distribution; finally, a decision rule. The details are as follows.

\subsection{Model for the target distributions}
\label{ssec:model_target_distributions}

GCCS aims to learn the most discriminative features and maximize the inter-class separability by finding a nonlinear projection of high-dimensional observations onto a lower-dimensional space. This is obtained by regularizing the latent space to $D$ different statistical distributions, where $D$ is the number of classes the data belongs to. Let us first define the desired target distribution $\mathbb{P}_i$ for class $\mathcal{C}_{i}$, $i=1,\ldots D$, as a $D$-variate Gaussian distribution, i.e. ${\mathbb{P}_i = \mathcal{N}(\muv_{Ti},\Sigmam_{T})}$,
with ${\muv_{Ti} = \mu_{T} \mathbf{e}_i}$ and ${\Sigmam_{T} = \sigma^2_{T} \mathbb{I}_D}$, where $\mathbf{e}_i$ is the $i$th standard unit vector and $\mathbb{I}_D$ is the $D\times D$ identity matrix. $\mu_{T}$ and $\sigma_{T}$ are user-defined parameters that are related to inter-class separation and are discussed later in the manuscript. Here, it should be noted that in order to have separable distributions we should have $\mu_T/\sigma_T > \sqrt{2D}$, otherwise as $D$ grows the classes will inevitably mix.

Since each distribution $\mathbb{P}_i$ has mean value proportional to $\mathbf{e}_i$, the statistical distributions  are centered on the vertices of a regular $(D-1)$-simplex at $\mu_T \mathbf{e}_i$, as shown in Fig. \ref{fig:RobustNet-arch}. 
This target model has several advantages. 
First, this choice guarantees that each class is at the same distance from all other classes. Due to the uniformity of the feature distributions in the latent space and the consequent lack of a short path, the attack strength must be much larger in order to generate a misclassification, leading to improved robustness. Moreover, since the distributions are Gaussian, the decision boundaries are straightforward to compute. This is in contrast with the typical behavior of neural networks, which tend to yield very complex boundaries, and it promotes accuracy as well as adversarial robustness.

\begin{table*}[t]
 \centering
\resizebox{\textwidth}{!}{
\begin{tabular}{ C{7cm}|  C{2cm}|  C{2cm} | C{2cm} | C{2cm} | C{3cm} | C{3cm} }
\hline
\noalign{\vskip 0.5mm}
\multirow{2}{*}{\textbf{Method}} & \textbf{MNIST} & \textbf{FMNIST} & \textbf{SVHN} & \textbf{CIFAR-10} & \textbf{CIFAR-10} & \textbf{CIFAR-100} \\ 
& \textbf{ResNet-18} & \textbf{ResNet-18} & \textbf{ResNet-18}  & \textbf{ResNet-18} & \textbf{Shake-Shake-96} & \textbf{Shake-Shake-112} \\ 
\Xhline{3\arrayrulewidth}
\noalign{\vskip 0.5mm}

GCCS - regular training & 99.58  & 92.69 & 94.20 & 82.97  & 96.19 & 76.53 \\

\textbf{GCCS - fine tuning} & \textbf{99.64 } & \textbf{93.83 } & \textbf{95.58 } & \textbf{81.52 }  & \textbf{97.06 } & \textbf{77.48 } \\

No Defense - cross-entropy & 99.35  & 91.91  & 94.12 & 78.59  & 95.78  & 76.30  \\ 

Jacobian Reg. - regular training \cite{jakubovitz2018improving} & 98.99  & 91.79  &  94.11  & 70.09  & - & -  \\ 

Jacobian Reg. - fine-tuning\cite{jakubovitz2018improving} & 98.53  & 92.43  & 93.54   & 82.09  & - & -  \\ 

Input Gradient Reg. - regular training \cite{ross2018improving} & 97.98  & 88.45  & 93.77  & 78.32  & 96.50  & 74.89  \\ 

Input Gradient Reg. - fine-tuning \cite{ross2018improving} & 99.11  & 92.55  & 93.17   & 76.15  & 96.90 & 75.68   \\ 

Cross Lipschitz regular training \cite{hein2017formal} & 96.78  & 92.54 &  91.42  & 80.10  & - & -  \\ 

Cross Lipschitz - fine-tuning \cite{hein2017formal} & 98.77  & 92.41  &  93.50  & 79.39  & - & -  \\ 

 \Xhline{3\arrayrulewidth}
  \end{tabular}}
\caption[]{Maximum test accuracy obtained through \textit{regular training} vs \textit{fine-tuning} over different benchmark datasets with different competing techniques in the case in which no adversarial attack is performed.}
\label{table:acc_comp}
\end{table*}

\subsection{Loss function}
\label{ssec:training}

In order to train the network, we need to introduce a loss function that allows us to minimize a suitable distance metric between the distributions of the output latent variables and the target distributions. 

Let us refer to the output of the encoding neural network as $\zv = H(\xv)$, where $[z_1, \dots, z_D] \in \mathbb{R}^D$ indicates the latent mapping, and $\xv \in \mathbb{R}^n$ denotes the input data belonging to $D$ different classes. The goal is to learn an encoding function of the input $\zv = H(\xv)$ such that ${\zv \sim \mathbb{P}_i}$ if ${\xv \in \mathcal{C}_{i}}$.

During the training phase, the network is given as input a batch of samples $\Xm \in \mathbb{R}^{b \times n}$, where $b$ is the batch size, and it computes the encoded outputs $\Zm \in \mathbb{R}^{b \times D}$. We are interested in their first and second order statistics, which can be estimated as sample mean $\muv_{Oi}$ and sample covariance $\Sigmam_{Oi}$ for each class.
Considering that the target statistics are known and the sample statistics for the batch have been computed, we can proceed to define a suitable loss to measure how far the distributions are from each other. More in detail, we employ the \textit{Kullback-Leibler} divergence (KL).

For the sample distribution of any class $\mathcal{C}_i$, the KL divergence with respect to the Gaussian target distribution can be written as:
\begin{equation}
\begin{split}
    \mathcal{L}_i = \log \frac{|\Sigmam_{T}|}{|\Sigmam_{Oi}|} - D + \mathrm{tr}(\Sigmam_{T}^{-1}\Sigmam_{Oi}) + \\ (\muv_{Ti}-\muv_{Oi})^\intercal \Sigmam_{T}^{-1}(\muv_{Ti}-\muv_{Oi} )  
\end{split}
\end{equation}
\label{eq:1}

\noindent We consider the cumulative loss $\mathcal{L} = \sum_{i=1}^{D}\mathcal{L}_i$. This loss reaches its minimum when the sample statistics of the $D$ encoded distributions match the target ones. However, for a small batch size, it can be difficult to control the behavior of the tails of the obtained distributions relying only on KL. Hence, we also consider the Kurtosis $\mathcal{K}_{i,j}$, \cite{joanes1998comparing} of the $j$th component of the $i$th target distribution, defined as  $\mathcal{K}_{i,j}= \left(\frac{z_{i,j}-\muv_{Oi,j}}{\sigma_{Oi,j}}\right)^4$.

In the case of multiple i.i.d. univariate normal distributions such as those we are enforcing at training, the target Kurtosis for each class is $\mathcal{K}_{i,j}=3$. This can be added to the cumulative loss, obtaining the loss ${\cal L}^{{\rm GCCS}}$ as follows: 

\begin{equation}
    {\cal L}^{{\rm GCCS}} = \sum_{i=1}^{D}\left[\mathcal{L}_i + \lambda(\mathcal{K}_i-3)\right],
\end{equation}
\label{eq:4}

\noindent where $K_i = 1/D\sum_j K_{i,j} $ and $\lambda$ determines the strength of the Kurtosis term and is set to $\lambda=0.2$.

\subsection{Decision Rule}
\label{ssec:test}

Once the preconditions are fulfilled, GCCS allows to define optimal decision boundaries in the resulting latent space. For the given target distributions, the optimal boundaries are obtained by partitioning the space into Voronoi regions such that all points in a region are closer to the respective centroid (the mean vector $\boldsymbol\mu_{Ti}$) than to any other centroid in the $(D-1)$-simplex. The resulting decision rule consists of computing the distance of the feature point from all centers and choose the class with the minimum distance. To determine which class a test image belongs to, the following decision rule is employed:

\begin{equation}
    \widehat{y}= \arg \max_i z_i,
\end{equation}
\label{eq:argmax}
\noindent which returns the index of the predicted class for the test image.

\section{Experiments}
\label{sec:experiments}

\subsection{Datasets and Training Parameters}
\label{sec:datasets}

The performance of classifiers trained using the GCCS loss was tested on MNIST \cite{lecun1998gradient}, FMNIST \cite{xiao2017fashion}, SVHN \cite{netzer2011reading}, CIFAR-10 and CIFAR-100 \cite{krizhevsky2009learning}.  For less complex datasets such as MNIST, FMNIST, and SVHN, the experiments were conducted using ResNet-18 \cite{he2016deep} as the feature extraction network. For the more challenging CIFAR-10 and CIFAR-100 datasets, the Shake-Shake-96 and Shake-Shake-112 \cite{gastaldi2017shake} regularization networks have been employed respectively, using a widen factor equal to \(6\) for the former and \(7\) for the latter. The encoder's last layer is followed by a fully-connected layer that outputs a vector with dimension \(D\).
We trained each network for a total of \(1800\) epochs. For better network convergence, we employed cosine learning rate decay \cite{gotmare2018closer} with an initial value of \(0.01\) as well as weight decay with a rate set to \(0.001\). Finally, dropout regularization \cite{srivastava2014dropout} with a \(0.8\) keep probability value was applied to all the fully connected layers in the network. 

\subsubsection{Target Distributions Parameters}
\label{ssec:target_distrib_parameters}

In this section, we perform an experiment to explore the behavior of the target distributions for different mean and variance values. Since we fix the mean $\mu_{T}$ and variance $\sigma_{T}$ values for the target distributions so that they are centered on the vertices of a regular $(D-1)$-simplex, the only parameter affecting our design is the $\mu_{T}/\sigma_{T}$ ratio, i.e., how far apart the distributions are with respect to the chosen variance. 

In this experiment we set $\sigma_T = 1$, so that the target distributions are $\mathbb{P}_i = \mathcal{N}(\mu_T \mathbf{e}_i,\mathbb{I}_D)$; then, we compute the classification accuracy as a function of $\mu_{T} \in [0.5,300]$. Fig. \ref{fig:parameters_select} shows the accuracy as a function of $\mu_T/\sigma_T$ for MNIST-10 dataset. It can be observed that in the $\mu_{T} \geq 20$ region the accuracy is even higher than that obtained with the traditional cross-entropy loss. 

In the following, we choose $\mu_T=70$ and $\sigma_T=1$. This choice ensures that we operate in that region, and also that the target distributions are sufficiently far apart from each other.

\begin{figure*}[t]
\centering
\begin{subfigure}{.19\textwidth}
  \centering
  \includegraphics[width=\linewidth]{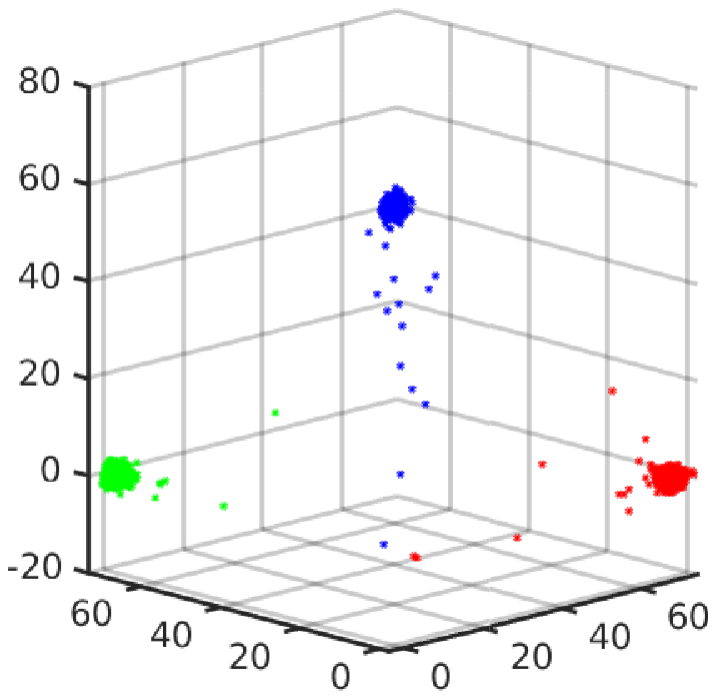}
  \caption[]{{GCCS}}
\end{subfigure}
\begin{subfigure}{.19\textwidth}
  \centering
  \includegraphics[width=\linewidth]{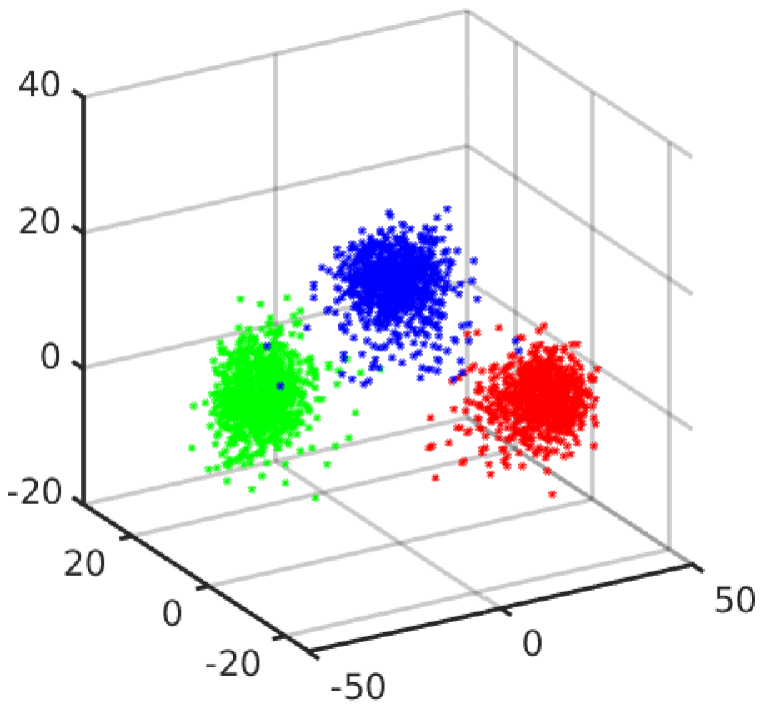}
  \caption[]{No defense}
\end{subfigure}
\begin{subfigure}{.19\textwidth}
  \centering
  \includegraphics[width=\linewidth]{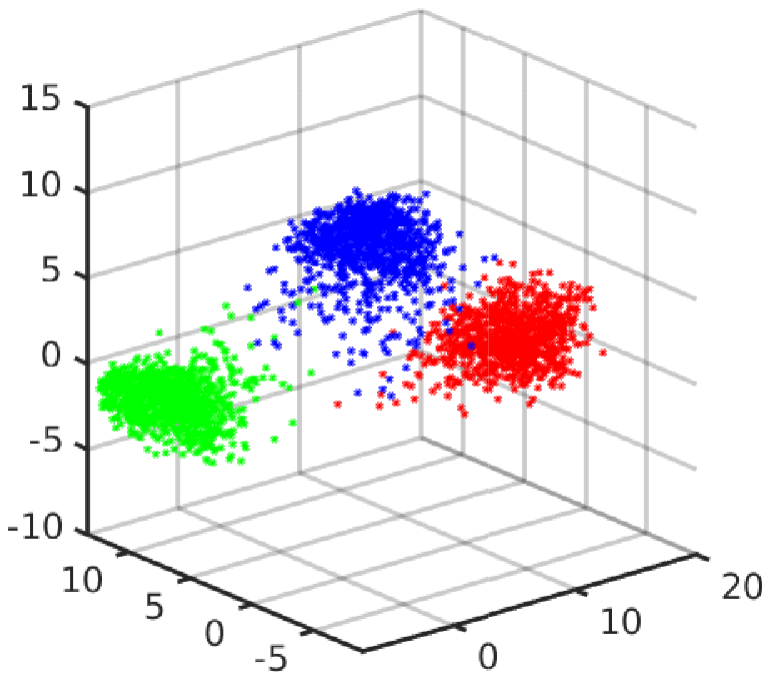}
  \caption[]{Jacobian Reg. }
\end{subfigure}
\begin{subfigure}{.19\textwidth}
  \centering
  \includegraphics[width=\linewidth]{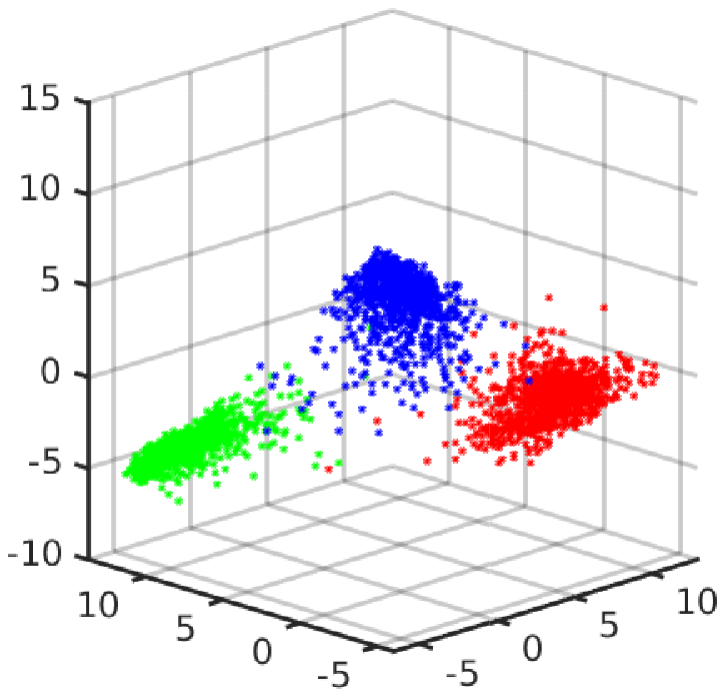}
  \caption[]{Input Reg.  }
\end{subfigure}
\begin{subfigure}{.19\textwidth}
  \centering
  \includegraphics[width=\linewidth]{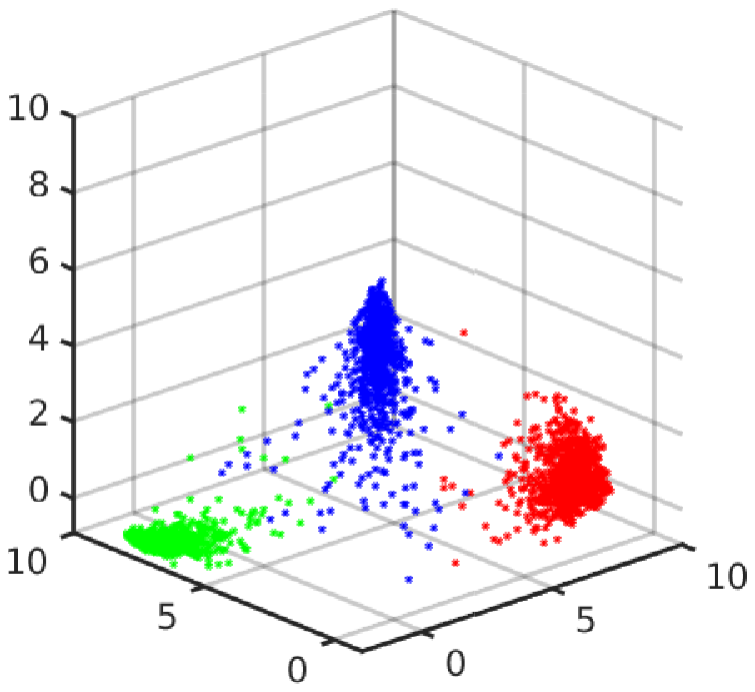}
  \caption[]{Cross Lipschitz}
\end{subfigure}
\begin{subfigure}{.19\textwidth}
  \centering
  \includegraphics[width=\linewidth]{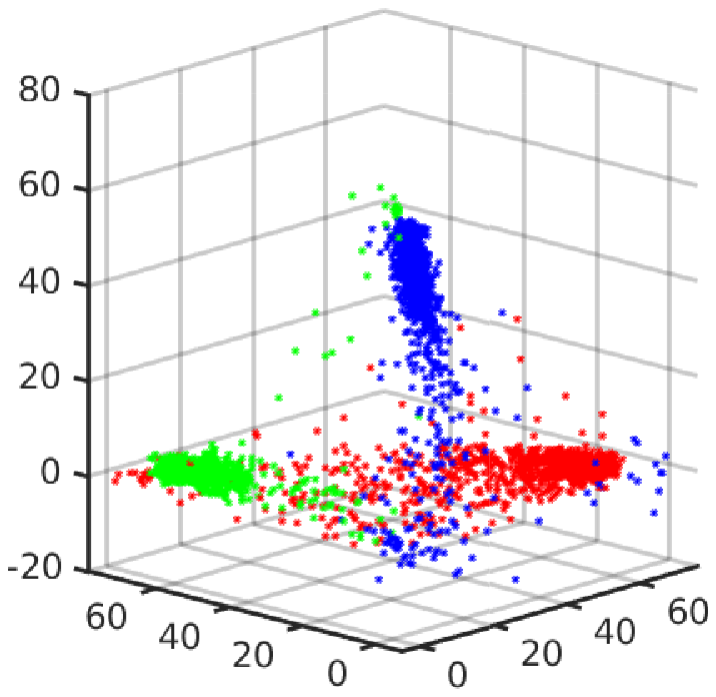}
  \caption[]{{GCCS}}
\end{subfigure}
\begin{subfigure}{.19\textwidth}
  \centering
  \includegraphics[width=\linewidth]{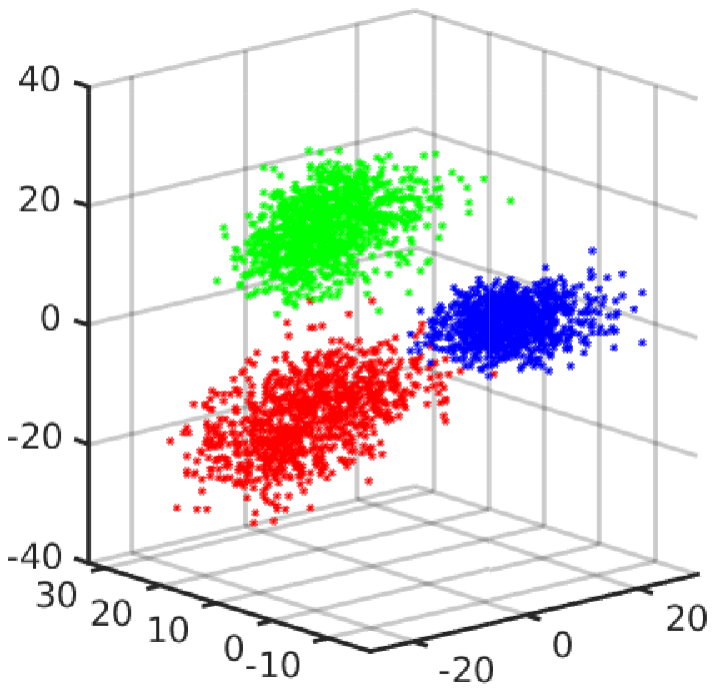}
  \caption[]{No defense}
\end{subfigure}
\begin{subfigure}{.19\textwidth}
  \centering
  \includegraphics[width=\linewidth]{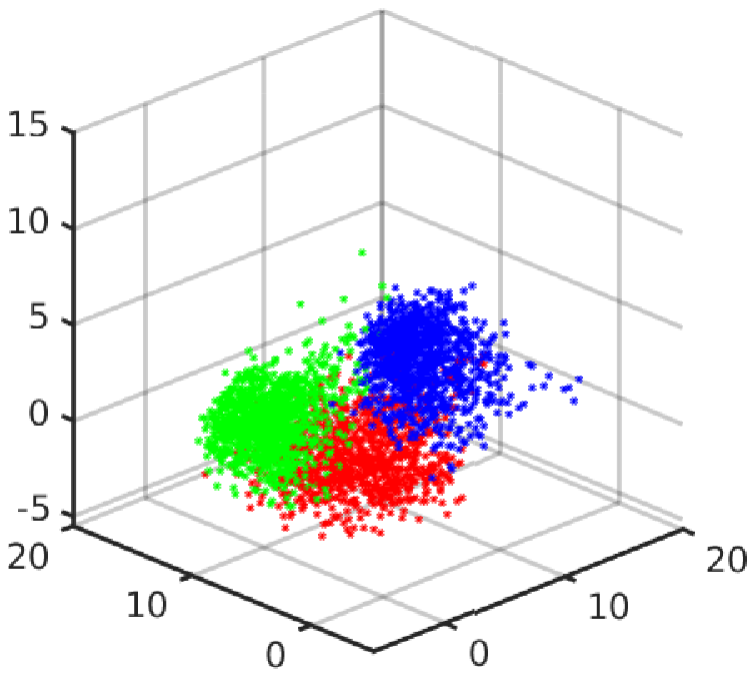}
  \caption[]{Jacobian Reg. }
\end{subfigure}
\begin{subfigure}{.19\textwidth}
  \centering
  \includegraphics[width=\linewidth]{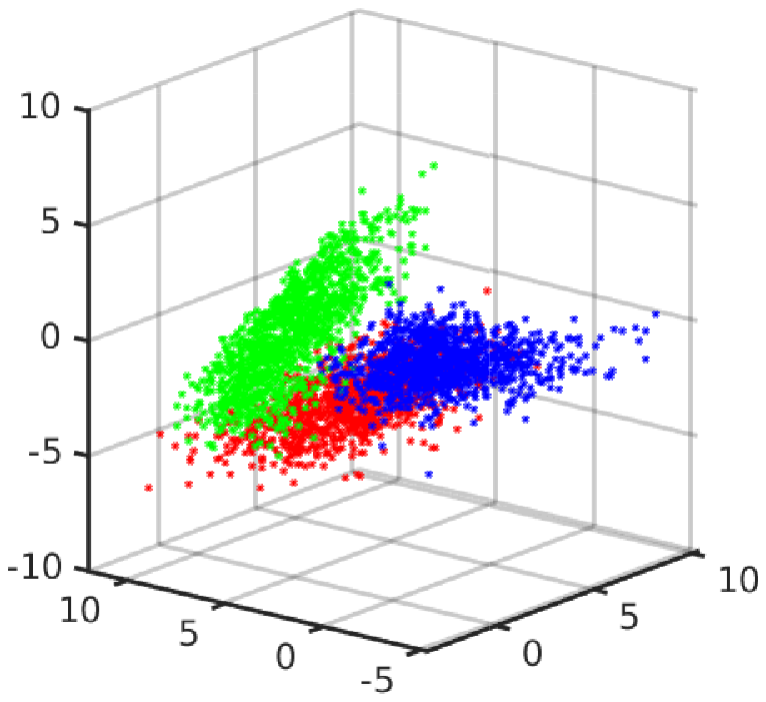}
  \caption[]{Input Reg.  }
\end{subfigure}
\begin{subfigure}{.19\textwidth}
  \centering
  \includegraphics[width=\linewidth]{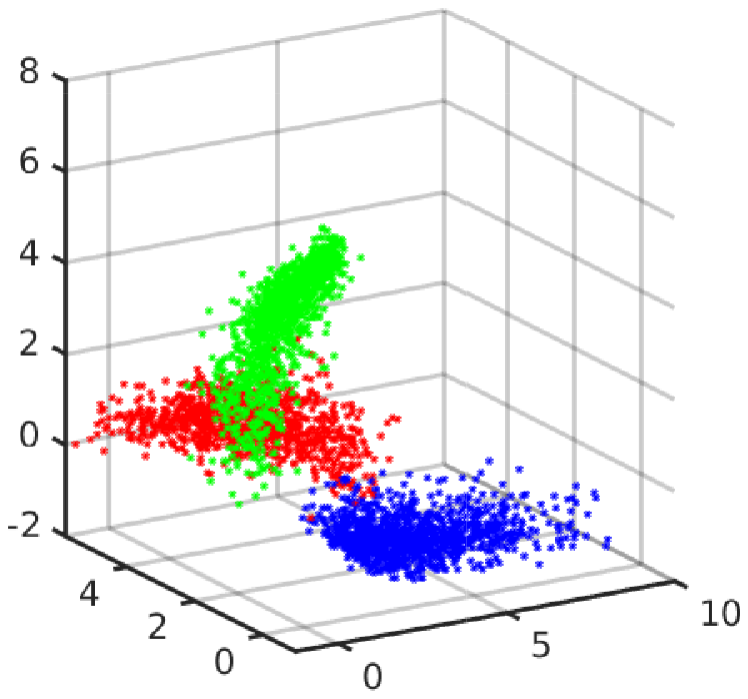}
  \caption[]{Cross Lipschitz}
\end{subfigure}

\caption[]{\textbf{(a-e)} Visual representation of latent space output distributions on MNIST for regular training in the case that no adversarial attack is applied. For better visualization of the separability, only three classes are shown, and an appropriate scale is used for each plot. (a) GCCS; (b) standard cross-entropy; (c) Jacobian Regularization \cite{jakubovitz2018improving}; (d) Input Gradient Regularization \cite{ross2018improving}; (e) Cross Lipschitz Regularization \cite{hein2017formal}. \textbf{(f-j)} Visual representation of latent space output distributions on MNIST for TGSM (5 steps, $\epsilon=2e^{-3}$) is applied. For better visualization, only three classes are shown. (f) GCCS; (g) standard cross-entropy; (h) Jacobian Regularization \cite{jakubovitz2018improving}; (i) Input Gradient Regularization \cite{ross2018improving}; (j) Cross Lipschitz Regularization \cite{hein2017formal}.}
\label{fig:mnist_3_original}
\vspace*{-0.5cm}
\end{figure*}

\subsection{Classification accuracy}
\label{sec:network_convergence}

As a first experiment, we compared the classification accuracy of GCCS with that obtained by an equivalent network trained with cross-entropy loss (no defense) and with state-of-the-art defense techniques such as Jacobian Regularization \cite{jakubovitz2018improving}, Input Gradient Regularization \cite{ross2018improving}, and Cross Lipschitz regularization \cite{hein2017formal}, in the case in which no adversarial attack is performed.
As shown in Table~\ref{table:acc_comp}, GCCS yields high classification accuracy both when the networks are trained from scratch (\textit{regular training}) and when they are first trained using regular cross-entropy loss and then fine-tuned with either GCCS loss or the other defense techniques (\textit{fine-tuning}). In particular, Table~\ref{table:acc_comp} shows that the proposed technique outperforms the standard cross-entropy loss and other existing approaches \cite{jakubovitz2018improving}, \cite{ross2018improving}, and \cite{hein2017formal} over the considered datasets. In more detail, it can be seen that other techniques generally cause a small decrease in classification accuracy with respect to the standard cross-entropy loss function, whereas GCCS provides an improvement in testing accuracy, especially for challenging datasets such as CIFAR-10 and CIFAR-100. 

The higher classification accuracy yielded by GCCS is due to the high separability of the target distributions in the latent space, as opposed to the other methods. To better highlight this, we refer to Fig.~\ref{fig:mnist_3_original} in which the output distributions for three different MNIST classes [0, 1 and 9] are reported. Looking at Fig.~\ref{fig:mnist_3_original}-a against Fig.~\ref{fig:mnist_3_original}-b, Fig.~\ref{fig:mnist_3_original}-c, Fig.~\ref{fig:mnist_3_original}-d, and Fig.~\ref{fig:mnist_3_original}-e, one can immediately observe that the output distributions of the three classes are less spread out and more separated than the other cases.

Also, Fig.~\ref{fig:mnist_3_original} shows that GCCS provides lighter distribution tails, compared to the other methods.

\subsection{Robustness Evaluation}
\label{sec:RobustNess_evl}

In this section, we evaluate how the classification accuracy of GCCS and the other competing techniques degrades under both targeted attacks (TGSM, JSMA) and non-targeted attack (PGD). The accuracy is evaluated as a function of a tunable parameter $\epsilon$ that indicates how strong the applied attack is. 

Namely, the noise vector $\nv$ added by the attack to the input signal $\xv$ satisfies $\lVert \nv\rVert_{\infty}/\lVert\xv\rVert_{\infty} \leq \epsilon$.


\begin{figure*}[t]
\centering
\begin{subfigure}{.24\linewidth}
  \centering
  \includegraphics[width=\linewidth]{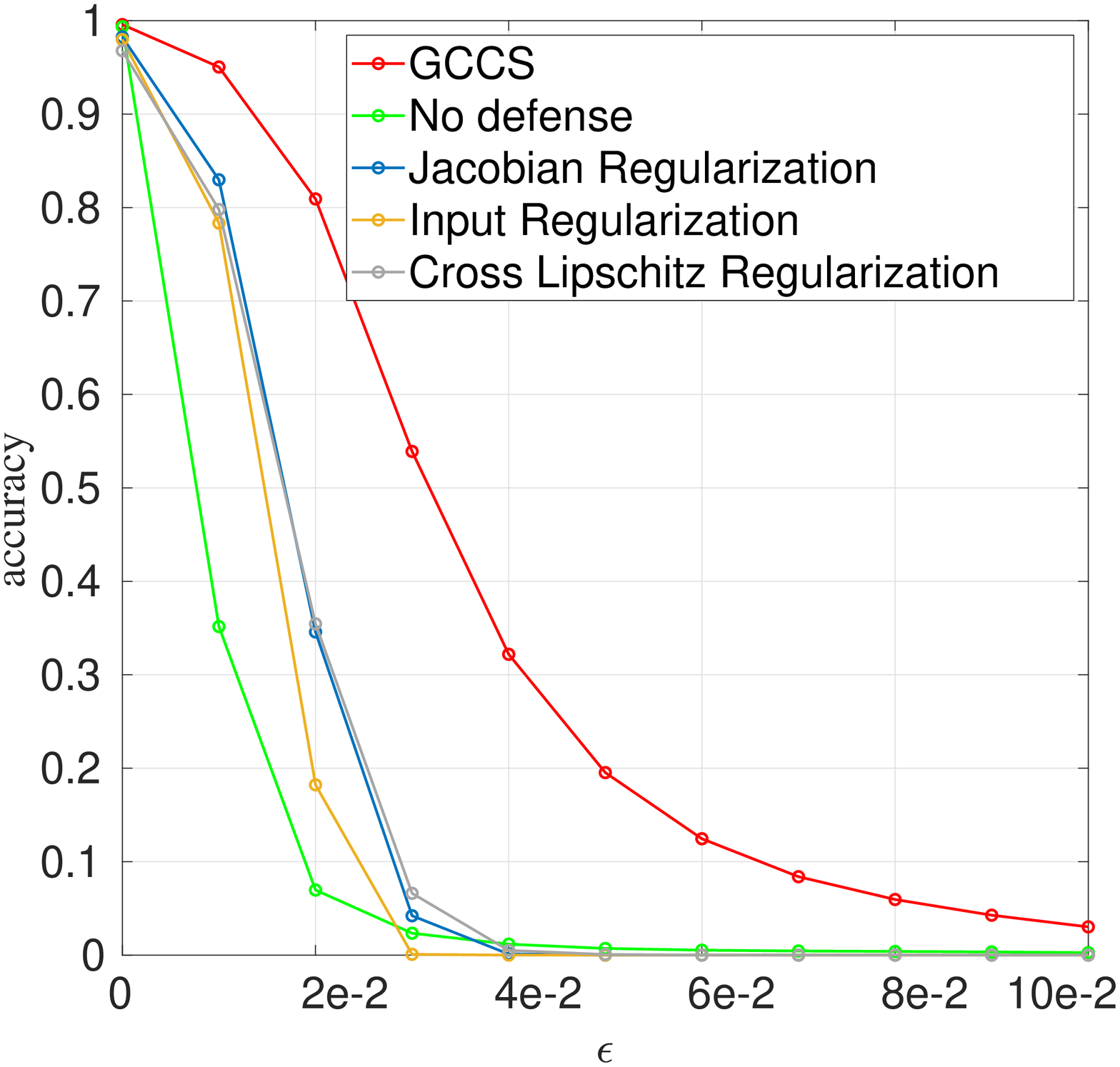}
  \caption[]{MNIST@ ResNet-18}
\end{subfigure}
\begin{subfigure}{.24\linewidth}
  \centering
  \includegraphics[width=\linewidth]{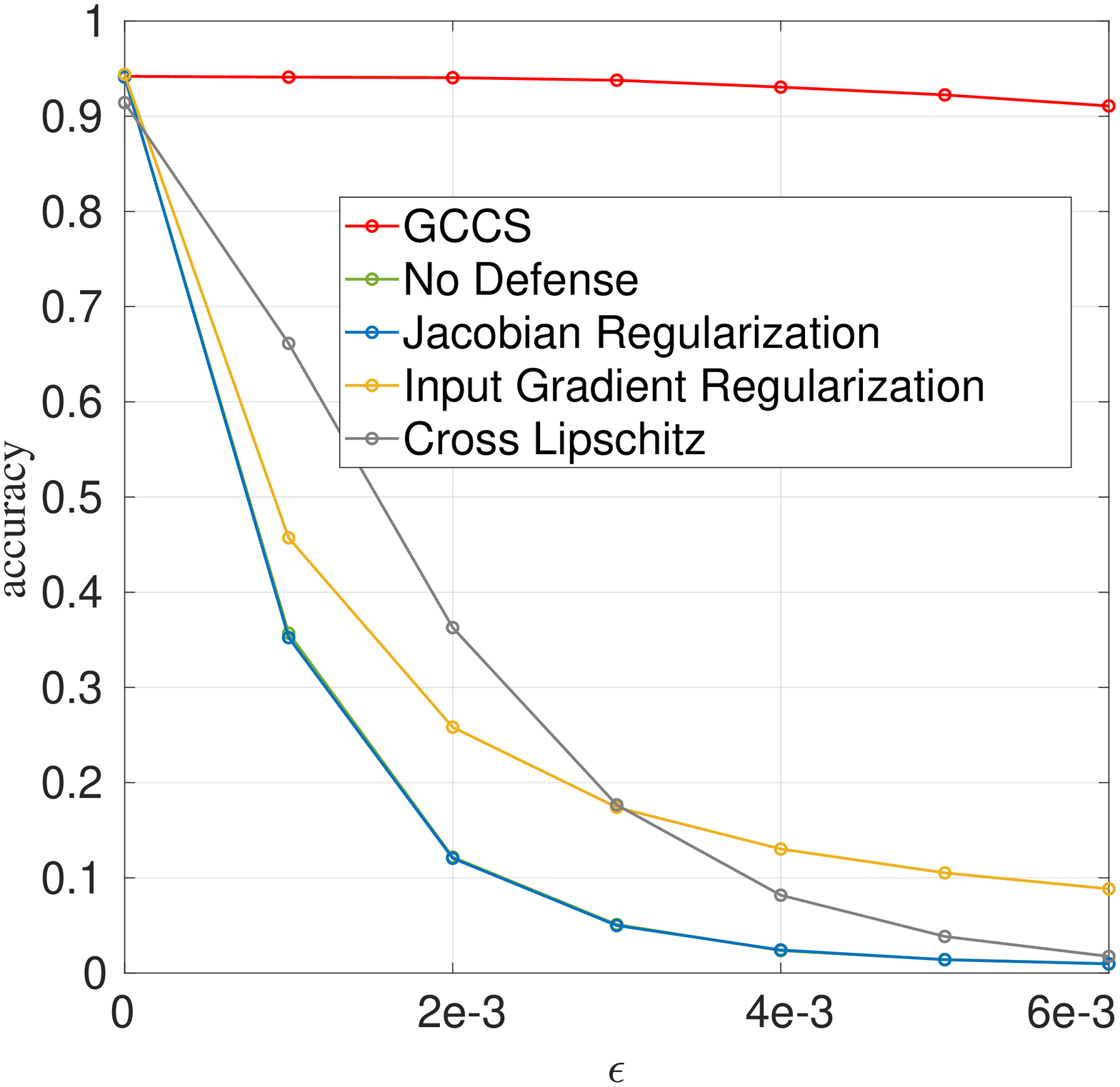}
  \caption[]{SVHN@ ResNet-18}
\end{subfigure}
\begin{subfigure}[htp]{.24\linewidth}
  \centering
  \includegraphics[width=\linewidth]{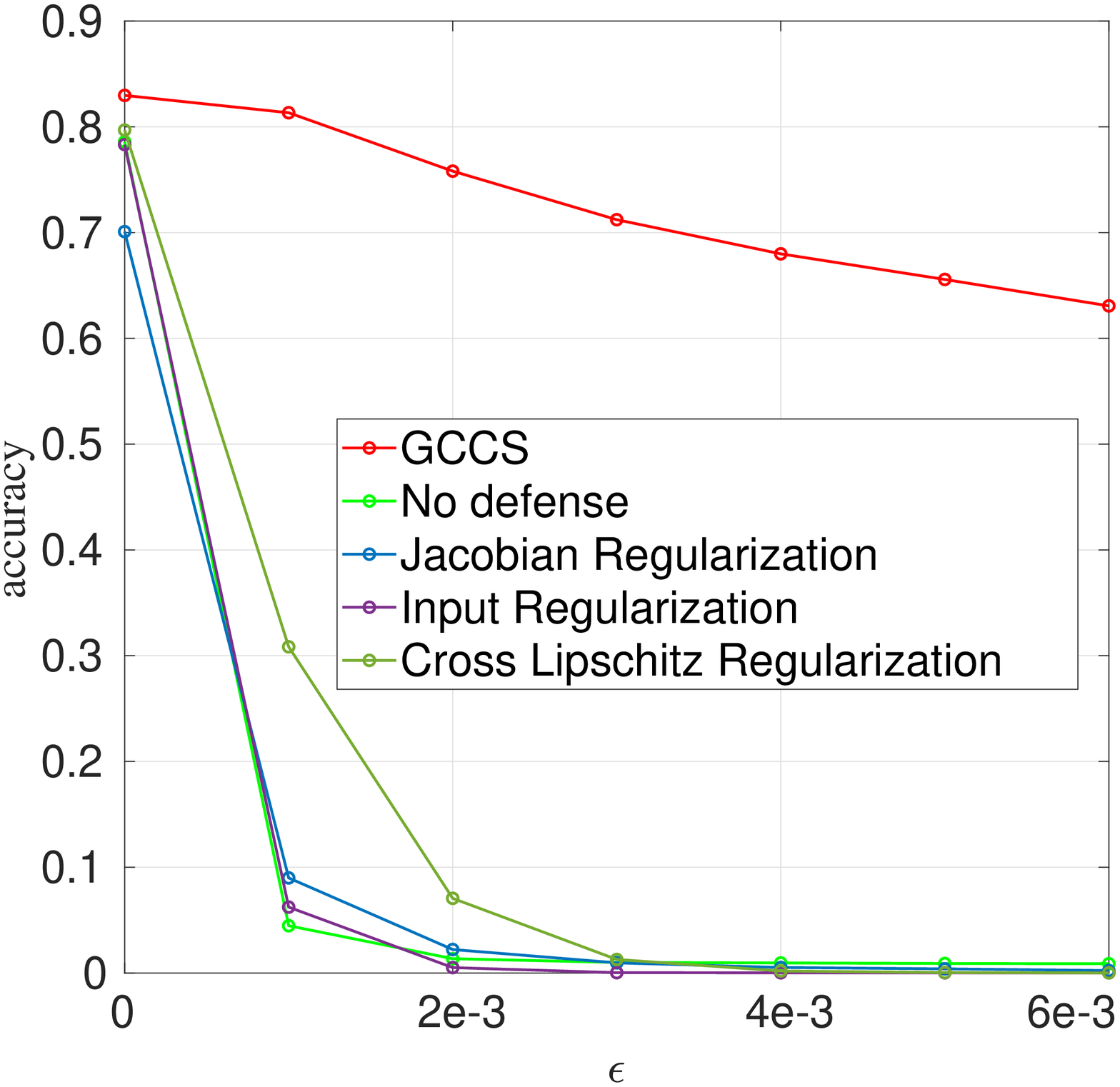}
  \caption[]{{Cifar10@ ResNet-18}}
\end{subfigure}
\begin{subfigure}[htp]{.24\linewidth}
  \centering
  \includegraphics[width=\linewidth]{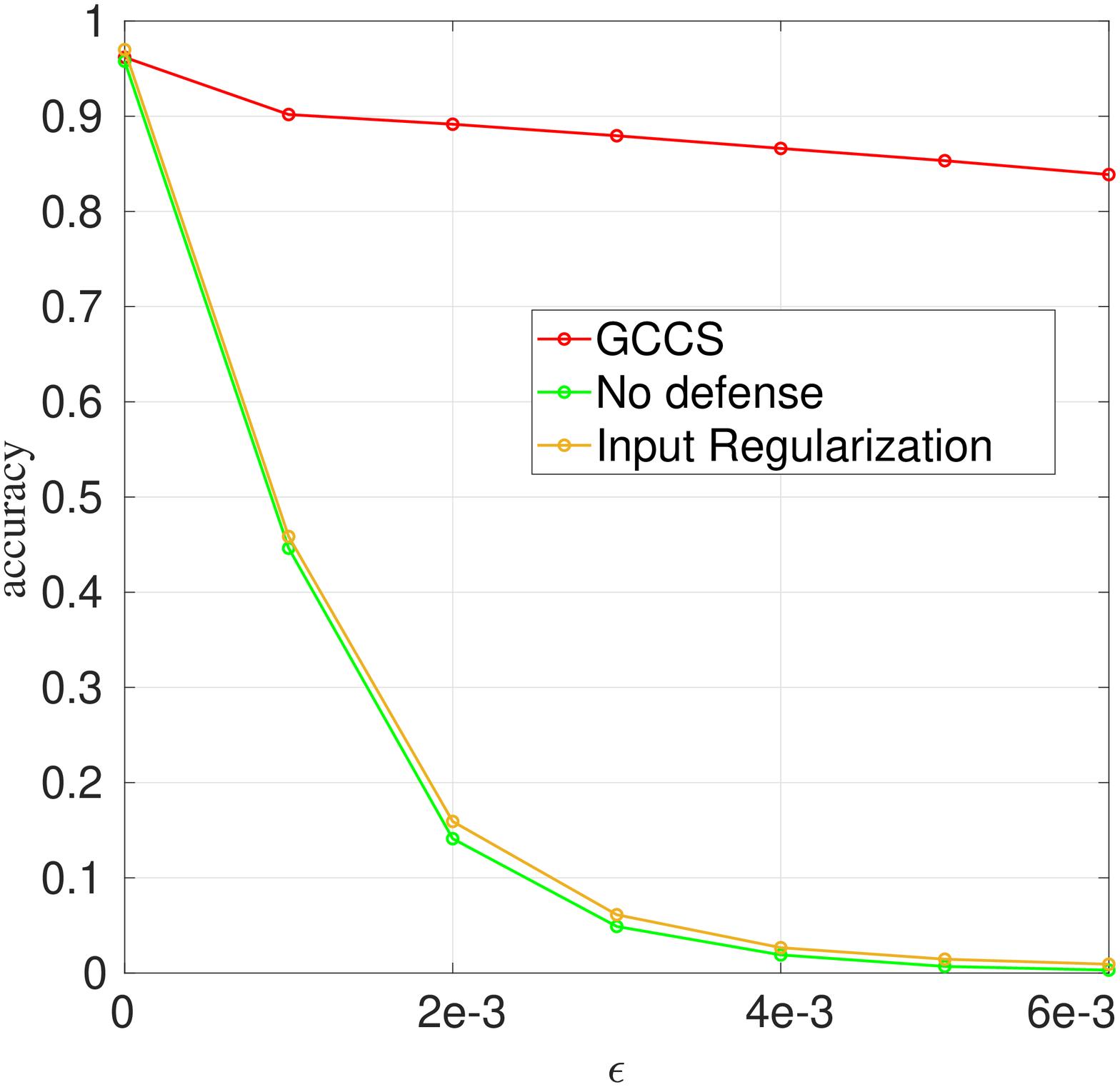}
  \caption[]{{Cifar10@ Shake-Shake}}
\end{subfigure}
\caption[]{Test accuracy for PGD (5 steps) attack on (a) ([MNIST, ResNet-18]); (b) ([SVHN, ResNet-18]); (c) ([CIFAR-10, ResNet-18]); (d) ([CIFAR-10, Shake-Shake-96]) for different values of $\epsilon$.}
\label{fig:PGD_evl}
\vspace*{-0.2cm}
\end{figure*}

\subsubsection{Non-targeted Attacks}
\label{ssec:nontargeted_attacks}

We start by evaluating the performance of all methods when subjected to the non-targeted PGD attack on the MNIST, SVHN, CIFAR-10, and CIFAR-100 datasets. Projected Gradient Descent (PGD) \cite{madry2017towards}, is an iterative version of FGSM in which noise is added in multiple steps. In particular, PGD is the strongest adversarial attack that exploits first-order local information about the trained model. In this work, for PGD we apply a \(5\)-iterations attack, i.e. PGD-\(5\) as done in \cite{shafahi2018adversarial, zheng2018pgd, davchev2019empirical}.

For MNIST, we set $0 \leq \epsilon \leq 10e^{-2}$, while for SVHN, CIFAR-10, and CIFAR-100 we set $0 \leq \epsilon \leq 6e^{-3}$, since MNIST is, in general, a less challenging dataset. As illustrated in Fig.~\ref{fig:PGD_evl}, GCCS outperforms by a large amount the competing approaches on all the considered datasets. Our approach proves to be much more robust than the others, especially for stronger attacks. The performance gap is particularly evident in the case of PGD, which is indeed the strongest adversarial attack utilizing the local first-order network information. 

\subsubsection{Targeted Attacks}
\label{ssec:targeted_attacks}

We also consider targeted adversarial attacks such as TGSM and JSMA.
Similarly to Sec.~\ref{ssec:nontargeted_attacks}, we present curves of the classification accuracy against the attack strength $\epsilon$.

\begin{figure*}[t]
\centering
\begin{subfigure}{.24\linewidth}
  \centering
  \includegraphics[width=\linewidth]{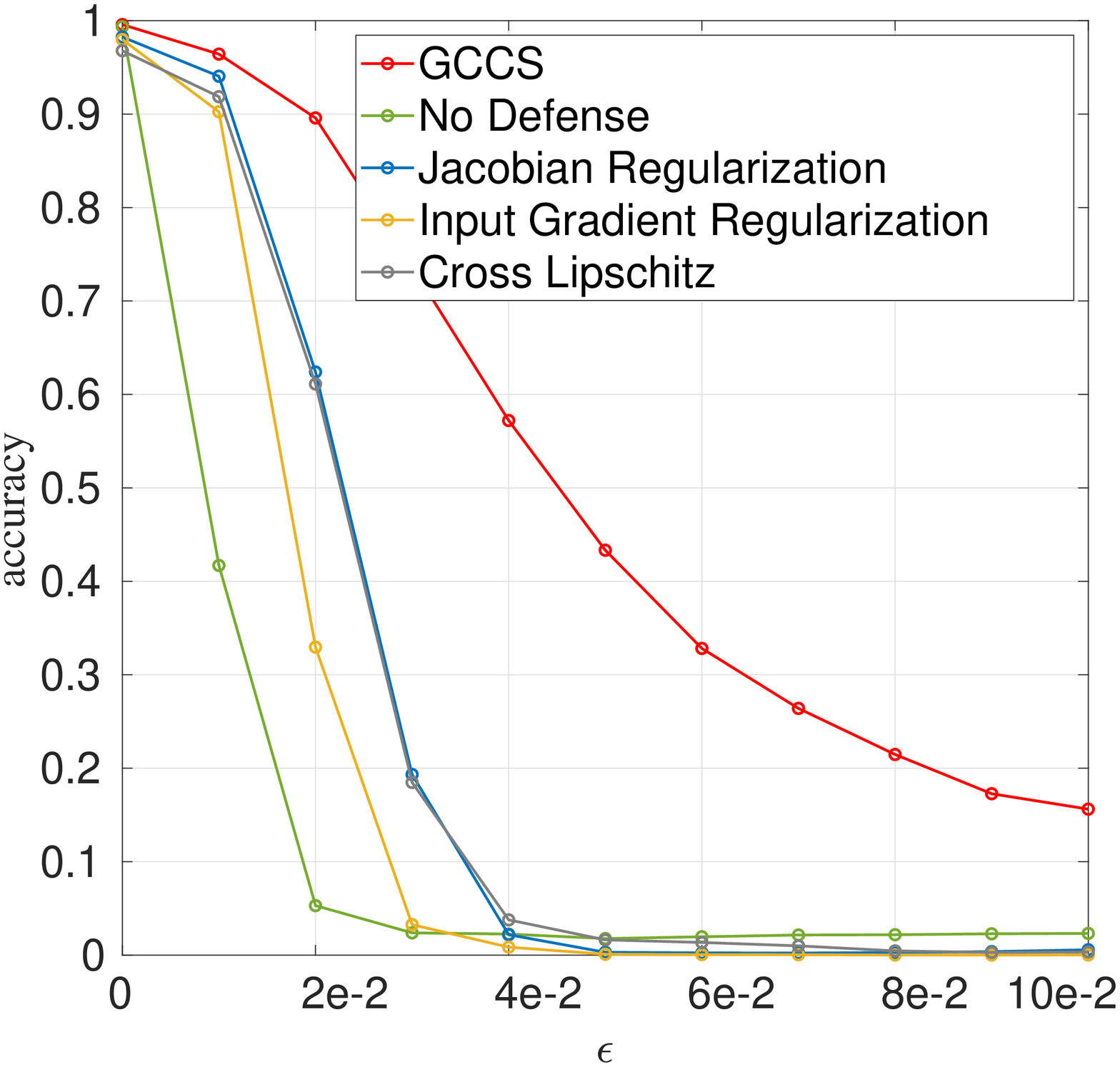}
  \caption[]{MNIST@ ResNet-18}
\end{subfigure}
\begin{subfigure}{.24\linewidth}
  \centering
  \includegraphics[width=\linewidth]{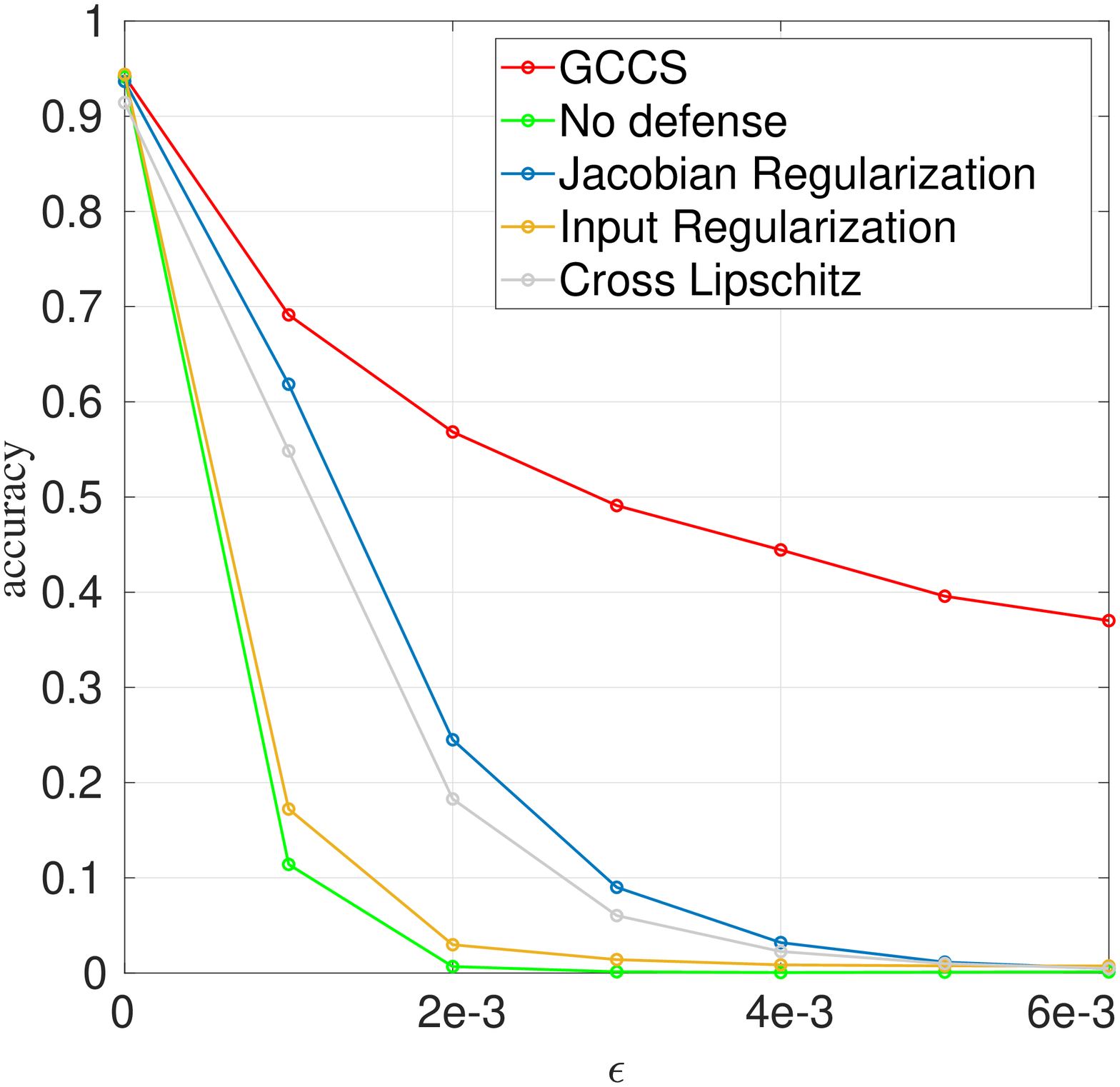}
  \caption[]{SVHN@ ResNet-18}
\end{subfigure}
\begin{subfigure}[htp]{.24\linewidth}
  \centering
  \includegraphics[width=\linewidth]{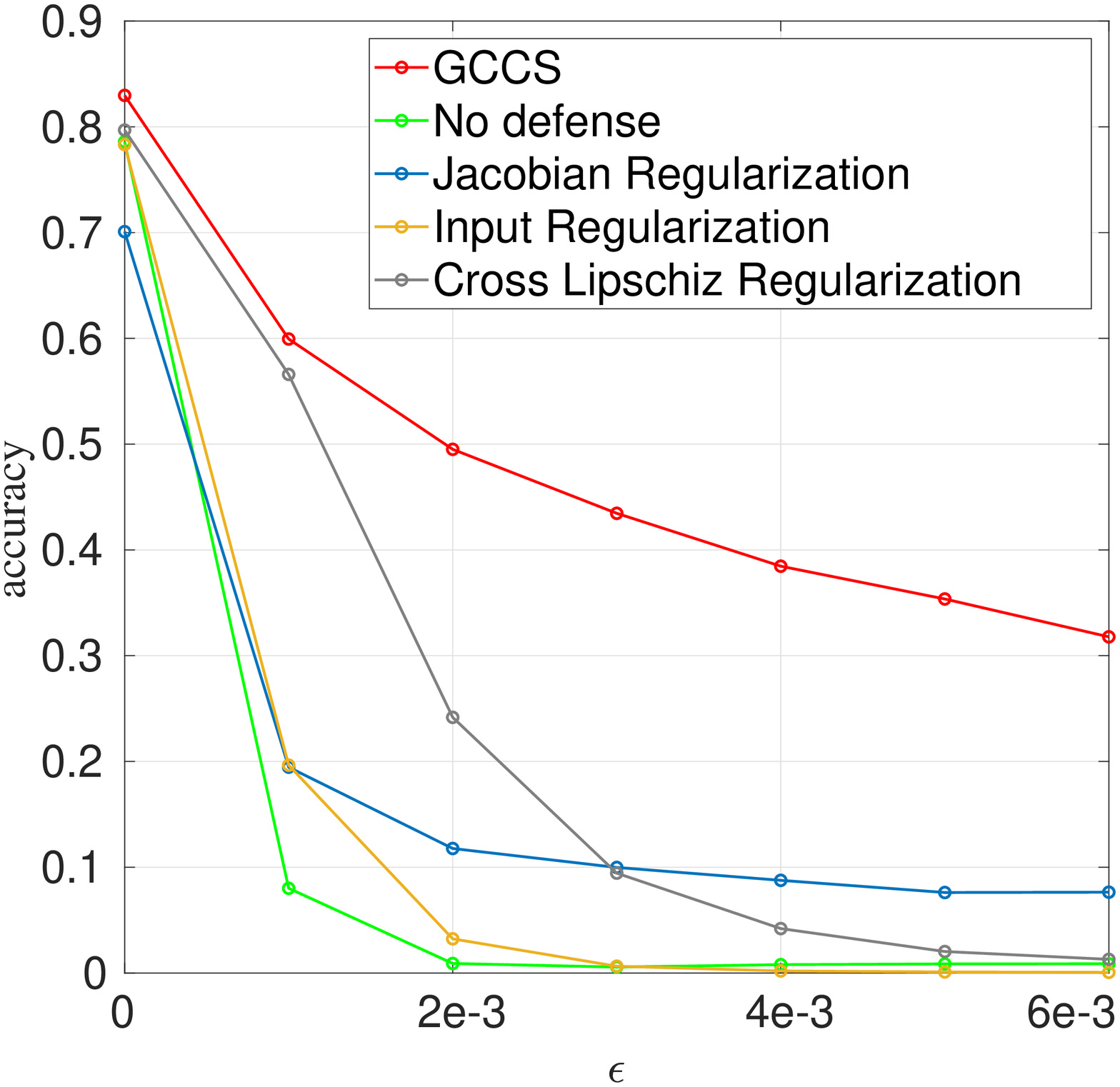}
  \caption[]{{Cifar10@ ResNet-18}}
\end{subfigure}
\begin{subfigure}[htp]{.24\linewidth}
  \centering
  \includegraphics[width=\linewidth]{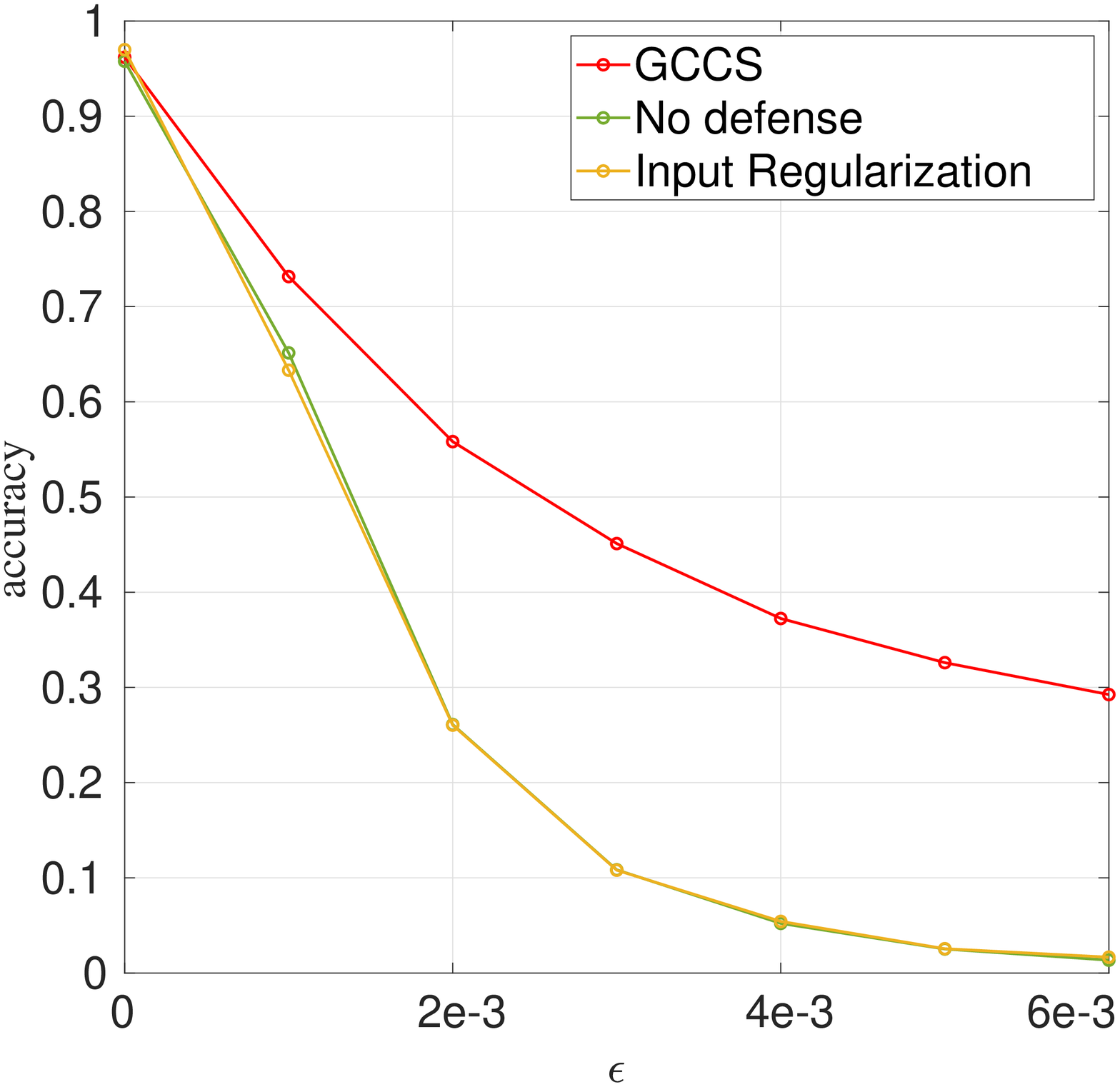}
  \caption[]{{Cifar10@ Shake-Shake}}
\end{subfigure}
\caption[]{Test accuracy when applying the TGSM attack (\(5\) steps) for (a) ([MNIST, ResNet-18]) ; (b) ([SVHN, ResNet-18]); (c) ([CIFAR-10, ResNet-18]) (d) ([CIFAR-10, Shake-Shake-96]), for different values of $\epsilon$.}
\label{fig:TGSM_evl}
\vspace*{-0.2cm}
\end{figure*}

\textbf{TGSM Attack}: In TGSM \cite{kurakin2016adversarial} the input samples are perturbed by adding noise in the direction of the negative gradient with respect to a selected target class. Fig.~\ref{fig:TGSM_evl} presents the results for TGSM-\(5\), a \(5\)-iterations TGSM attack, over the MNIST, SVHN, CIFAR-10, and CIFAR-100 datasets. In this attack, the targeted output class is $y_{l+1}$ when the true class is $y_l$.

\begin{figure*}[t]
\centering
\begin{subfigure}{.24\linewidth}
  \centering
  \includegraphics[width=\linewidth]{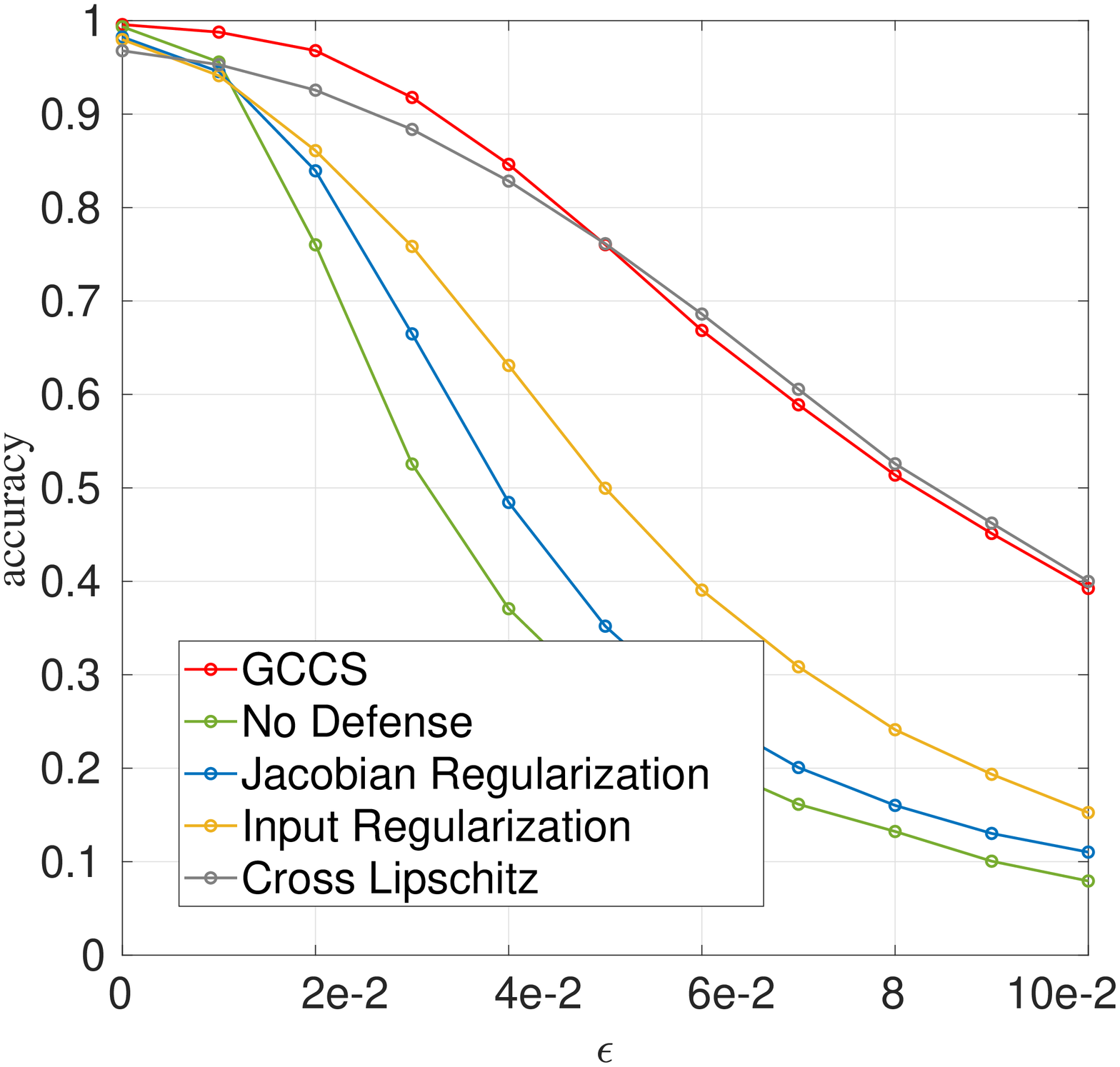}
  \caption[]{MNIST@ ResNet-18}
\end{subfigure}
\begin{subfigure}{.24\linewidth}
  \centering
  \includegraphics[width=\linewidth]{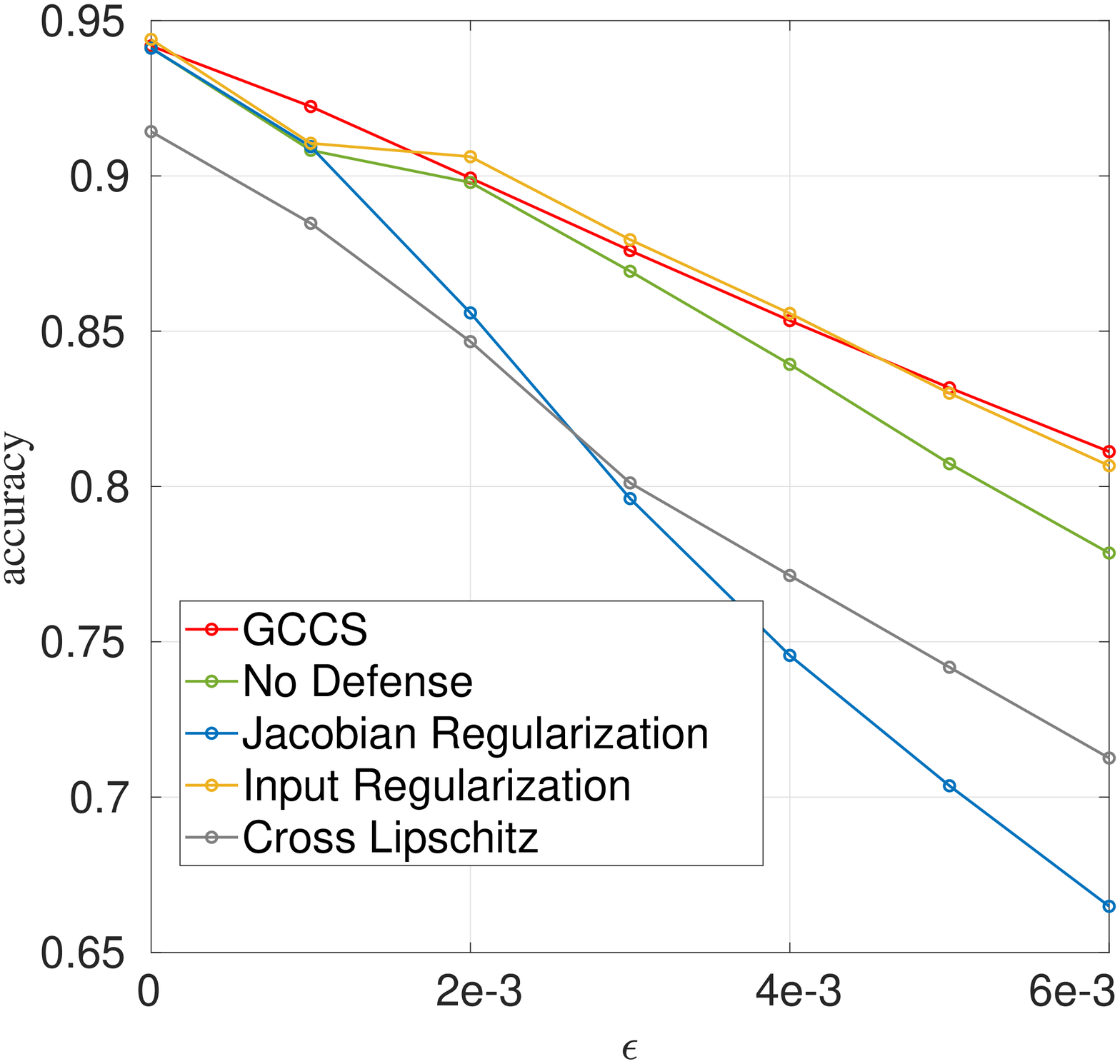}
  \caption[]{SVHN@ ResNet-18}
\end{subfigure}
\begin{subfigure}[htp]{.24\linewidth}
  \centering
  \includegraphics[width=\linewidth]{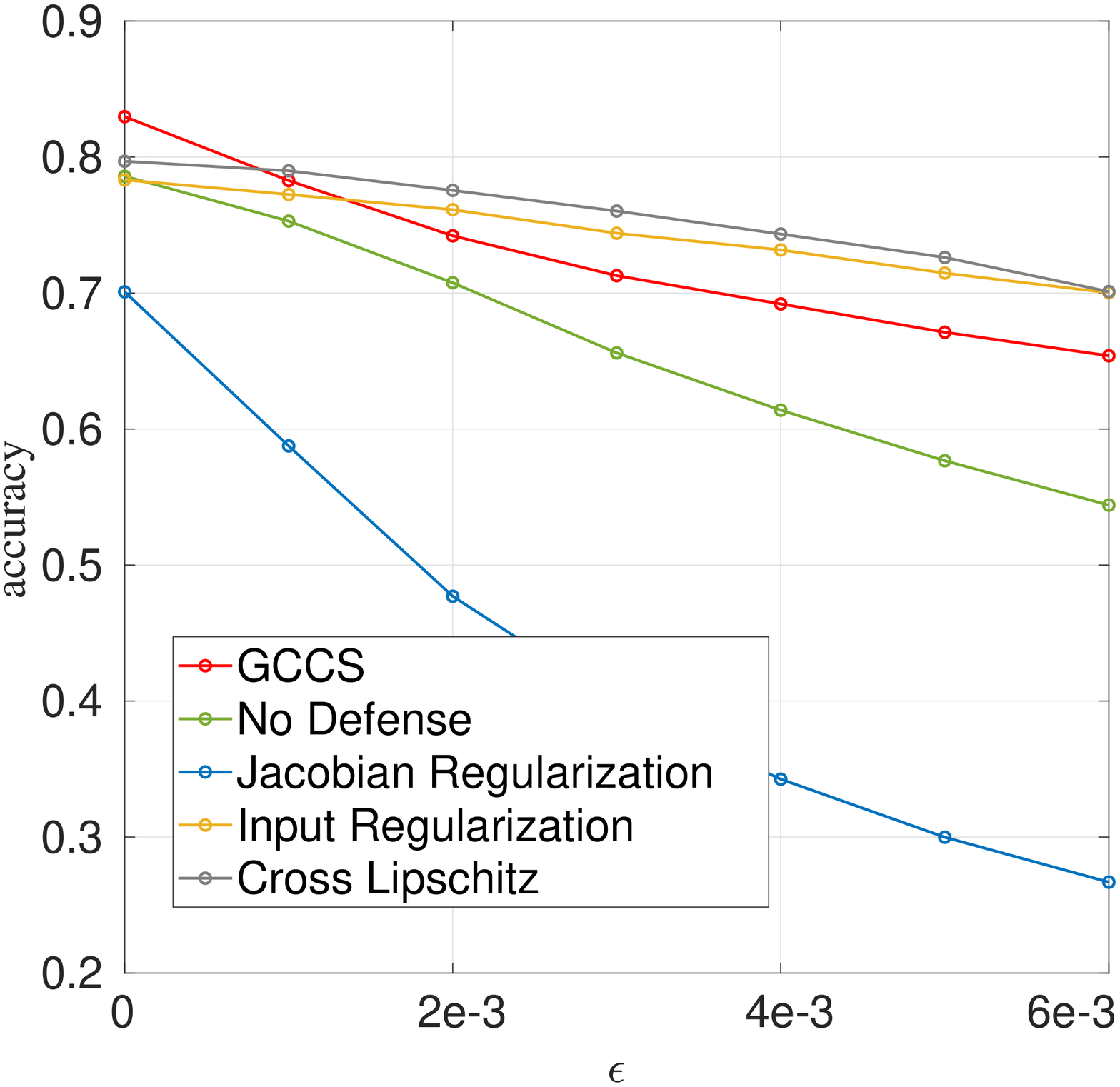}
  \caption[]{{Cifar10@ ResNet-18}}
\end{subfigure}
\begin{subfigure}[htp]{.24\linewidth}
  \centering
  \includegraphics[width=\linewidth]{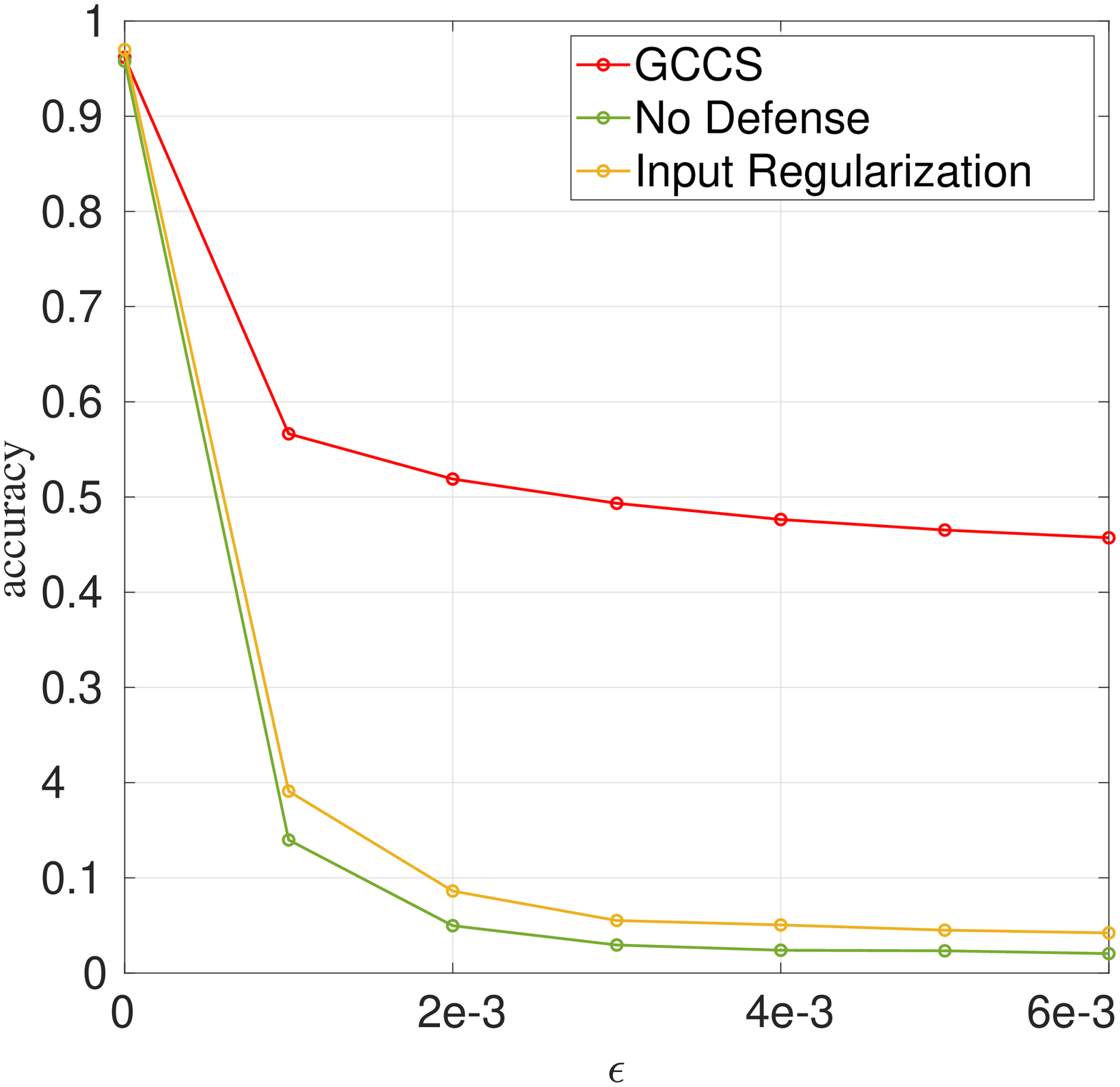}
  \caption[]{{Cifar10@ Shake-Shake}}
\end{subfigure}
\caption[]{Test accuracy when applying the JSMA attack (200 steps, 1 pixel) on (a) ([MNIST, ResNet-18]); (b) ([SVHN, ResNet-18]); (c) ([CIFAR-10, ResNet-18]); (d) ([CIFAR-10, Shake-Shake-96]), for different values of $\epsilon$.}
\label{fig:JSMA_evl}
\vspace*{-0.5cm}
\end{figure*}

It can be observed from Fig.~\ref{fig:TGSM_evl} that GCCS yields significantly higher performance compared to the other methods, throughout different datasets and with different attack strength $\epsilon$. In order to gain a better understanding of why the proposed method works much better than the others, in Fig.~\ref{fig:mnist_3_original}
we show a visual representation of the target distributions in the latent space {\em after} the TGSM-5 attack $\epsilon=2e^{-3}$ has been performed. 

Fig.~\ref{fig:mnist_3_original}-g shows clearly the effectiveness of the attack when no defense mechanism is employed, in the sense that the output distributions are shifted so as to replace the output distribution of the next class. 
Fig.~\ref{fig:mnist_3_original}-h, Fig.~\ref{fig:mnist_3_original}-i, and Fig.~\ref{fig:mnist_3_original}-j report the output distributions under TGSM in the case of Jacobian, Input Gradient, and Cross-Lipschitz regularization respectively, showing that, despite the defense mechanism, the distributions still tend to move their position in the latent space towards the adjacent classes, causing a very important drop in classification accuracy as seen in Fig.~\ref{fig:TGSM_evl}. In the GCCS case instead (Fig.~\ref{fig:mnist_3_original}-f), even if the tails of the output distributions become heavier, their positions are not swapped with the neighboring classes, allowing for better separability and hence improved classification accuracy and robustness.

\textbf{JSMA Attack}: The other targeted attack we consider is JSMA \cite{papernot2016limitations}, which consists in iteratively computing the Jacobian matrix of the network function to form a saliency map; this map is used at every iteration to choose which pixels to tamper with so that the likelihood of changing the output class towards a selected one is increased. In our case, we consider JSMA-\(200\) with a \(1\)-pixel saliency map. Similarly to the TGSM case, Fig.~\ref{fig:JSMA_evl} shows the classification accuracy for increasing attack strength \(\epsilon\). The proposed method confirms its robustness even to JSMA attack, achieving better robustness than other methods especially on the challenging CIFAR-10 dataset.

\section{Conclusions}
\label{sec:conclusion}

We have presented an approach that goes beyond cross-entropy, employing a loss function that promotes class separability and robustness by learning a mapping of the decision variables onto Gaussian distributions. Our work was motivated by the idea that mapping the centroids of the distributions on the vertices of a simplex could lead to the uniformity of the feature distributions in the latent space and the lack of a short path towards a neighboring decision region. Experiments on different multi-class datasets show excellent performance of the classifiers trained using the GCCS loss both in terms of accuracy and robustness of the classifier against adversarial attacks, outperforming existing state-of-the-art methods, both when used to train a network from scratch and when applied as a fine-tuning step on pre-trained networks. 
The performance is analyzed both for targeted and non-targeted adversarial attacks. We have shown that regularizing the latent space onto target distributions significantly increases the robustness against adversarial perturbations. Indeed, an analysis of the distributions in the latent space for the proposed GCCS method shows that the different classes tend to remain separated even in the presence of targeted attacks, whereas a similar attack strength invariably mixes the distributions achieved by competing methods.

\section{Acknowledgment}
This work results from the research cooperation with Sony R\&D Center Europe Stuttgart Laboratory 1.

\medskip


\bibliographystyle{unsrt}
{\small

\bibliography{references}
}

\end{document}